%% file: main.tex
\documentclass[a4paper, 10pt, conference]{ieeeconf}      % Use this line for a4
\usepackage{FG2021}
\FGfinalcopy % *** Uncomment this line for the final submission

\usepackage{times}
\usepackage{epsfig}
\usepackage{graphicx}
\usepackage{amsmath}
\usepackage{bm}
\usepackage{amssymb}
\usepackage{subcaption}
\usepackage[pagebackref=true,breaklinks=true,colorlinks,bookmarks=false]{hyperref}
\usepackage{booktabs}
\usepackage{tensor}
\usepackage{enumerate}

\usepackage[accsupp]{axessibility} % Improves PDF readability for those with disabilities. 

\IEEEoverridecommandlockouts                              % This command is only
\def\FGPaperID{84} % *** Enter the FG2021 Paper ID here

\title{\LARGE \bf
Tensor-based Subspace Factorization for StyleGAN
}

\author{\parbox{16cm}{\centering
    {\large  René Haas, Stella Graßhof and Sami S. Brandt}\\
    {\normalsize
    IT University of Copenhagen, Copenhagen, Denmark\\}}
}

\newcommand\R{\mathbb{R}}
\newcommand{\norm}[1]{\ensuremath{\left|\left|#1\right|\right|}}
\newcommand{\onenorm}[1]{\ensuremath{\norm{#1}_1}}
\renewcommand{\vec}[1]{\mathbf{#1}}
\newcommand{\matr}[1]{\mathbf{#1}}

\newcommand{\T}{\mathrm{T}}

\begin{document}
%%%%%%%%%%%%%%
% COPYRIGHT NOTICE - Uncomment correct version below
%
% The notices are from the FG 2021 LOA 
%
% Active is the "Others" option - see Case #4 in the instructions posted at: http://iab-rubric.org/fg2021
%
%%%%%%%%%%%%%%

% Case #1: For papers in which all authors are employed by the US government, the copyright notice is: 
%\IEEEoverridecommandlockouts\pubid{\makebox[\columnwidth]{U.S. Government work not protected by U.S. copyright \hfill}
%\hspace{\columnsep}\makebox[\columnwidth]{ }}

% Case #2: For papers in which all authors are employed by a Crown government (UK, Canada, and Australia), the copyright notice is:
%\IEEEoverridecommandlockouts\pubid{\makebox[\columnwidth]{978-1-6654-3176-7/21/\$31.00~\copyright{}2021 Crown \hfill}
%\hspace{\columnsep}\makebox[\columnwidth]{ }}

% Case #3: For papers in which all authors are employed by the European Union, the copyright notice is:
%\IEEEoverridecommandlockouts\pubid{\makebox[\columnwidth]{978-1-6654-3176-7/21/\$31.00~\copyright{}2021 European Union \hfill}
%\hspace{\columnsep}\makebox[\columnwidth]{ }}

% Case #4: For all other papers the copyright notice is:
\IEEEoverridecommandlockouts\pubid{\makebox[\columnwidth]{978-1-6654-3176-7/21/\$31.00~\copyright{}2021 IEEE \hfill}
\hspace{\columnsep}\makebox[\columnwidth]{ }}

\ifFGfinal
\thispagestyle{empty}
\pagestyle{empty}
\else
\author{Anonymous FG2021 submission\\ Paper ID \FGPaperID \\}
\pagestyle{plain}
\fi
\maketitle

% %%%%%%%%%%%%%%%%%%%%%%%%%%%%%%%%%%%%%%%%%%%%%%%%%%%%%%%%%%%%%%%%%%%%%%%%%%%%%%%%
% \begin{abstract}
% abstract
% \end{abstract}
\input{TensorGAN}
%%%%%%%%%%%%%%%%%%%%%%%%%%%%%%%%%%%%%%%%%%%%%%%%%%%%%%%%%%%%%%%%%%%%%%%%%%%%%%%%

{\small
\bibliographystyle{ieee}
%\bibliography{egbib}
\bibliography{references}
}

\end{document}

%% file: TensorGAN.tex
\begin{abstract}
\noindent
In this paper, we propose $\tau$GAN a tensor-based method for modeling the latent space of generative models.
The objective is to identify semantic directions in latent space. 
To this end, we propose to fit a multilinear tensor model on a structured facial expression database, which is initially embedded into latent space. 
We validate our approach on StyleGAN trained on FFHQ using BU-3DFE as a structured facial expression database. 
We show how the parameters of the multilinear tensor model can be approximated by Alternating Least Squares. 
Further, we introduce a stacked style-separated tensor model, defined as an ensemble of style-specific models to integrate our approach with the extended latent space of StyleGAN. 
We show that taking the individual styles of the extended latent space into account leads to higher model flexibility and lower reconstruction error.  
Finally, we do several experiments comparing our approach to former work on both GANs and multilinear models. 
Concretely, we analyze the expression subspace and find that the expression trajectories meet at an apathetic face that is consistent with earlier work. 
We also show that by changing the pose of a person, the generated image from our approach is closer to the ground truth than results from two competing approaches. 
\noindent
\end{abstract}

\section{Introduction}\label{sec:introduction}
In this paper, we propose a novel framework for finding semantic directions in the latent space of Generative Adversarial Networks (GANs) \cite{Goodfellow2014gan}. 
GANs have, since their proposal, emerged as one of the most dominant approaches for unsupervised representation learning in Computer Vision and beyond \cite{alea2020}. 

% Intro to GANs
Architecturally GANs refer to the simultaneous training of two neural networks: a \emph{generator} and a \emph{discriminator}.
The generator produces images by sampling from its latent space, while the discriminator, a binary classifier, tries to discriminate the generated images from the training images.
The goal of training is to reach the equilibrium of the min-max game between the two adversaries, such that neither can improve by changing the parameter values. 
At equilibrium, the discriminator can be discarded, and the generator can then be used to produce new data by sampling from the latent distribution.
The new data points follow the same statistics as the training data but are not contained in it. 
Modern state-of-the-art GAN variations have borrowed from the Style-transfer literature \cite{Huang2017StyleTransfer, Park2018StyleAttentional} to disentangle the latent space and synthesize high-quality face images. 
Work by \cite{Karras2019StyleGAN, Karras2019StyleGAN2}, and most recently \cite{Karras2020StyleGANADA}, showed how to a train state-of-the-art StyleGAN model, even in cases of limited data.

% Semantic directions in GANS
A recent goal has been to find semantically interpretable directions in GAN latent spaces, and
several approaches for \emph{semantic face editing} have been proposed. 
Semantic face editing refers to the ability to change various semantic attributes, such as identity, expression, and rotation, gender, of the generated images. 
Early work used an information criterion (InfoGAN) \cite{Chen2016InfoGAN} to determine semantic directions. However, as pointed out in \cite{Deng2020DiscoFaceGAN}, there is no guarantee that the latent codes produced by this method are semantically meaningful. 
Additional unsupervised approaches for finding semantic directions in StyleGAN include Principal Component Analysis (PCA) on sampled latent codes \cite{Harkonen2020GANSpace} and the closed-form factorization suggested by \cite{Shen2020ClosedForm}.

A recent approach for finding semantic directions in StyleGAN in a supervised fashion is to train binary linear classifiers (SVMs) to detect single binary semantic attributes such as smile vs. no smile, male vs. female, glasses vs. no glasses. For a given semantic attribute, the semantic direction could then be defined as the normal to the supporting hyper-planes of the trained SVM \cite{Shen2020interfacegan}. 

\begin{figure}[tb]
\centering
\includegraphics[width=0.95\linewidth,trim={1cm 0 0 0},clip]{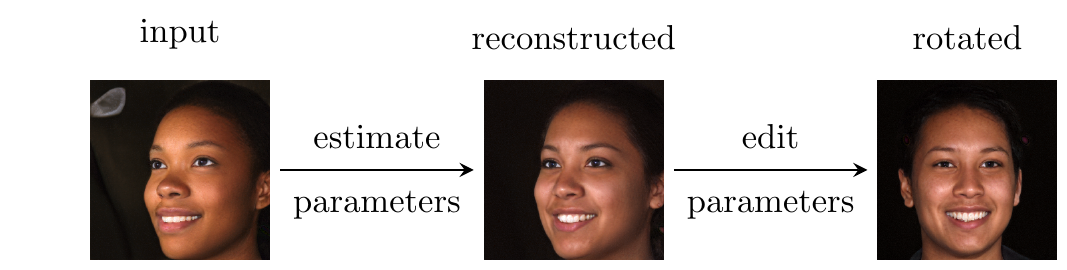} 
\caption{Overview of the proposed approach.} 
\label{fig1:diagram}
\end{figure}

% Multilinear models in the literature
In the literature, a wide collection of \emph{multilinear} methods have been proposed to model and analyze faces and expressions.  Early, PCA or dictionary-based 3D Morphable Models (3DMM) \cite{Blanz1999,Ferrari2017} capture the variation in shape and texture of neutral 3D faces. Recently 3DMMs have also been used to make semantic edits to images generated by StyleGAN \cite{Tewari2020Stylerig}.
More recently, factorization methods, based on higher-order data representations, were introduced with the benefit of better disentanglement of dimensions, such as person-specific shape and expression, when compared to
matrix methods \cite{tensorface, Vlasic2005}. 
These models were built on the Higher-Order Singular Value Decomposition (HOSVD) to factorize the data, and have successfully been used to model faces, their 3D reconstruction, as well as in transferring expressions \cite{Brunton2014,facewarehouse}. 
Moreover, in \cite{apathy,grasshof2020} a HOSVD tensor model was constructed from the Binghamton 3D facial expression database (BU-3DFE) \cite{bu3dfe}, which revealed a practically planar expression subspace, in which the six basic emotions form one-dimensional affine subspaces \cite{apathy}. 
These six lines intersect in a common vertex, the origin of expressions, which surprisingly does not represent the neutral face, but an extrapolated expression referred to as \emph{apathetic}. 

% Our contribution
The main novelty of this work is to use a multilinear face model to analyze the latent space of GANs. 
More specifically, we propose to use the HOSVD to factorize the latent space into semantically meaningful linear subspaces that yield a multilinear tensor model. 
Given an input image, we estimate the model parameters to approximate the input, and then change one attribute, such as rotation, as illustrated in Fig.~{\ref{fig1:diagram}}. 
% In other words, we propose a novel technique for semantic face editing.

\noindent
The main contributions of this paper are as follows:
\begin{itemize}%[noitemsep]
    \item We propose a novel method for semantic face editing with StyleGAN. 
    \item We propose a method to estimate model parameters and present reasonable regularization, enabling stable parameter transfer.
    \item We show that expression trajectories intersect at a unique point, corresponding to the origin of expressions, which differs from the neutral face confirming the earlier findings \cite{apathy,grasshof2020} based on BU-3DFE.
    \item We propose an extended model, based on style separation, which leads to greater model flexibility and lower reconstruction error for independent test images.
\end{itemize}

% Paper Organization
The paper is organized as follows: 
In Sec.~\ref{sec:stylegan} we will review the architecture and outline the process on how to embed reference images into the latent space of StyleGAN. 
In Sec.~\ref{sec:tensormodel} we present our Tensor-Based GAN model which we build "on top of" the StyleGAN latent space. Here we will also elaborate on how we can approximate model parameters for a given latent vector. 
Experiments and results of our proposed approach are presented in Sec.~\ref{sec:results}, followed by a summary and conclusion in Sec.~\ref{sec:conclusion}.

\section{StyleGAN}\label{sec:stylegan}
In this section, we will review the StyleGAN architecture and explain how to embed reference images into the latent space of the pre-trained models released by Nvidia \cite{Karras2019StyleGAN, Karras2019StyleGAN2}.

\subsection{StyleGAN Architecture}
\begin{figure}[tb]
\includegraphics[width=0.99\linewidth,trim={1cm 0 0 0 },clip]{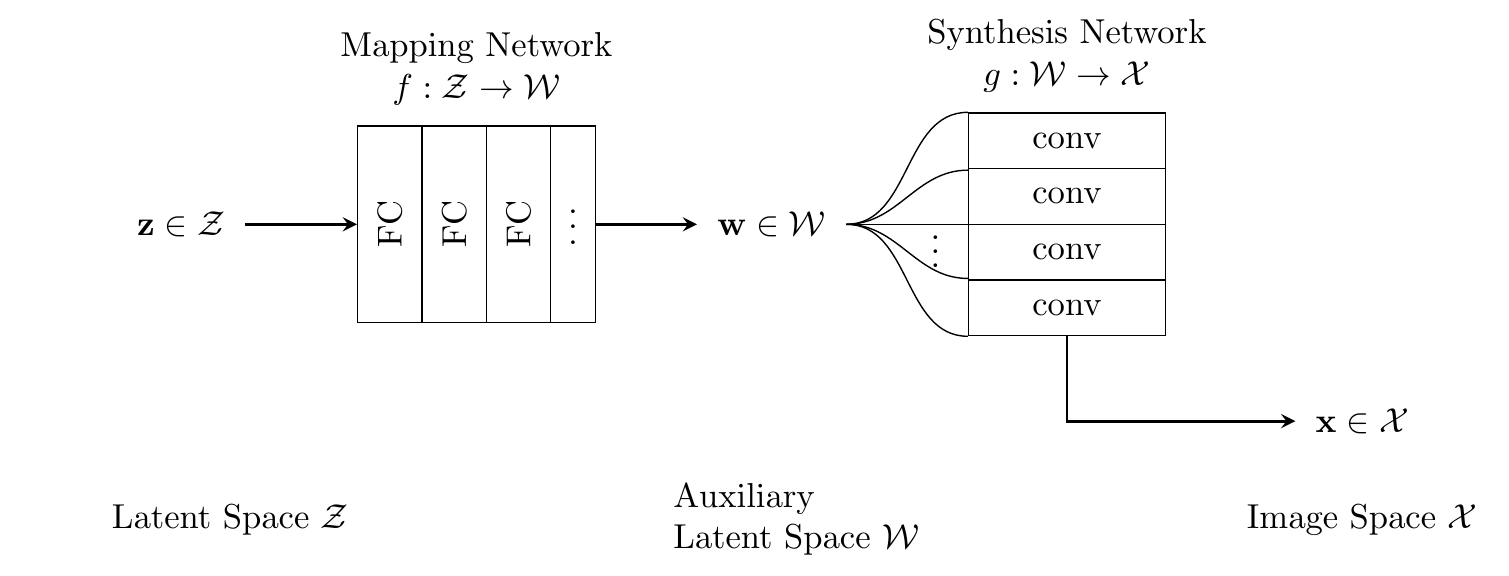}
\caption{Architecture of the StyleGAN generator.}
\label{fig:stylegan12}
\end{figure}

The StyleGAN \emph{generator} $G:\mathcal{Z}\to \mathcal{X}$, where $G = g\circ f$, is composed of two networks, the \emph{mapping network} $f:\mathcal{Z}\to\mathcal{W}$ and the \emph{synthesis network} $g:\mathcal{W}\to\mathcal{X}$, see Fig.~\ref{fig:stylegan12}.
The mapping network $f$, maps the latent vector $\vec{z}\in \mathcal{Z}$ onto the auxiliary latent space $\mathcal{W}$ to the vector $\mathbf{w}=f(\mathbf{z})$ while the synthesis network $g:\mathcal{W}\to\mathcal{X}$ maps the vector $\mathbf{w}\in\mathcal{W}$ to the final output image $\mathbf{x}\in\mathcal{X}$ in image space. 
The full generator $G$ thus maps the latent vector $\mathbf{z}$ to an image $\mathbf{x}$. 
The notation used in this paper is summarized in Tab.~\ref{tab:notation}.

\begin{figure}[tb]
\centering
\includegraphics[width=0.99\linewidth,trim={0.8cm 0 0 0 0},clip]{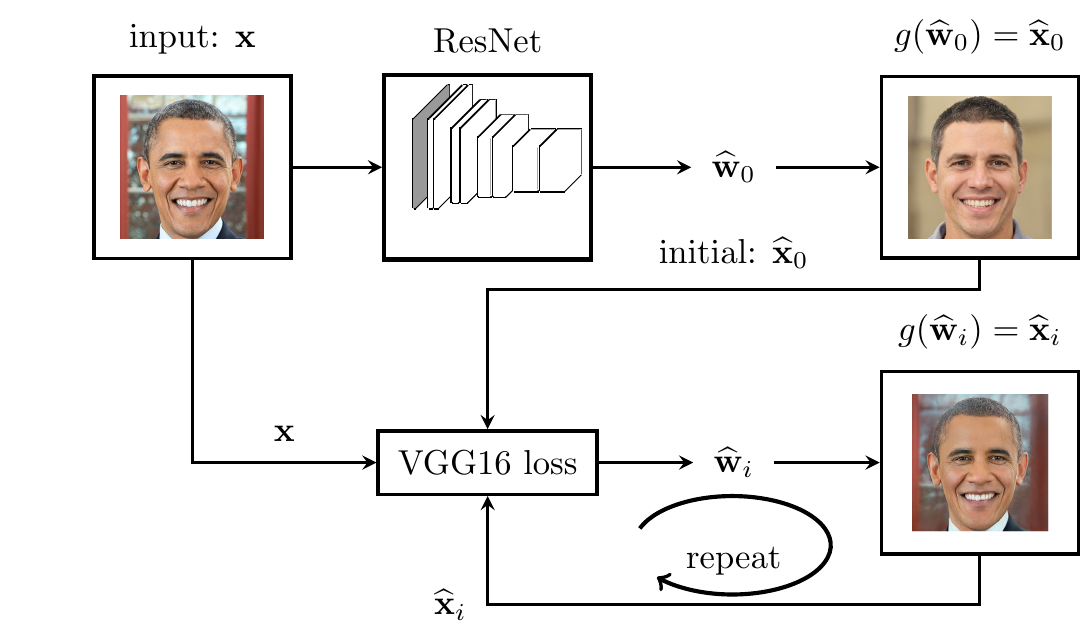} 
\caption{Diagram illustrating image embedding into the auxiliary latent space $\mathcal{W}$.}
\label{fig:embedding_pipeline}
\end{figure} 

\subsection{Generator Inversion}\label{subsec:generator_inversion}
GANs do not include an encoder as part of their architecture.  
Therefore, a goal in GAN research has been to find a method for finding a latent code that produces an image as close as possible to a given reference image, which we refer to as \emph{embedding} an image into the latent space. 
The problem can be considered as inverting the synthesis network $g^{-1}:\mathcal{X}\to \mathcal{W}$
%, and has been addressed in %has been considered by
\cite{abdal2019image2stylegan,puzer} while inverting $G$, and thereby embedding into $\mathcal{Z}$ space, has been investigated in \cite{Karras2019StyleGAN2}.
Contemporary techniques for $\mathcal{W}$ space embedding, i.e. finding $g^{-1}$, use a VGG network \cite{vgg16}.
Our approach for embedding onto the auxiliary latent space $\mathcal{W}$ is illustrated in Fig.~\ref{fig:embedding_pipeline}. 
The inverse generator $G^{-1}: \mathcal{X} \to \mathcal{Z}$ yields the latent vector $\mathbf{z}=G^{-1}(\mathbf{x})$ with $ G^{-1} = f^{-1} \circ g^{-1}$ for the input image $\mathbf{x}$. 

The initial estimate for the auxiliary latent vector for a given reference image is computed as follows. 
We use the pre-trained weights of StyleGAN \cite{Karras2019StyleGAN} and the recently revised StyleGAN2 \cite{Karras2019StyleGAN2} architecture.
Then, as proposed in \cite{pbaylies}, we train a ResNet \cite{He2016DeepRL} in a supervised setting using synthetic StyleGAN data to approximate $g^{-1}$ that yields the initial estimate $\hat{\mathbf{w}}_0$ for the latent vector. 
The refinement for the auxiliary latent vector is computed by first using the VGG16 network \cite{vgg16}, pre-trained on ImageNet database, and then removing the classification layer, hence the truncated network produces a high dimensional feature vector for a given input image, as described in \cite{Zhang2018unreasonable}.
Since the trained generator is fully differentiable, the loss can be calculated in VGG space and gradients back-propagated through the generator, hence we can iteratively update the latent code. 
This approach is also used in \cite{puzer}.
We also found that using the ResNet estimate as initialization for the VGG optimization process, leads to faster convergence than not using ResNet initialization.  

\section{Multilinear Model}\label{sec:tensormodel}
\begin{figure}[tb]
    \centering
    \includegraphics[width=0.99\linewidth]{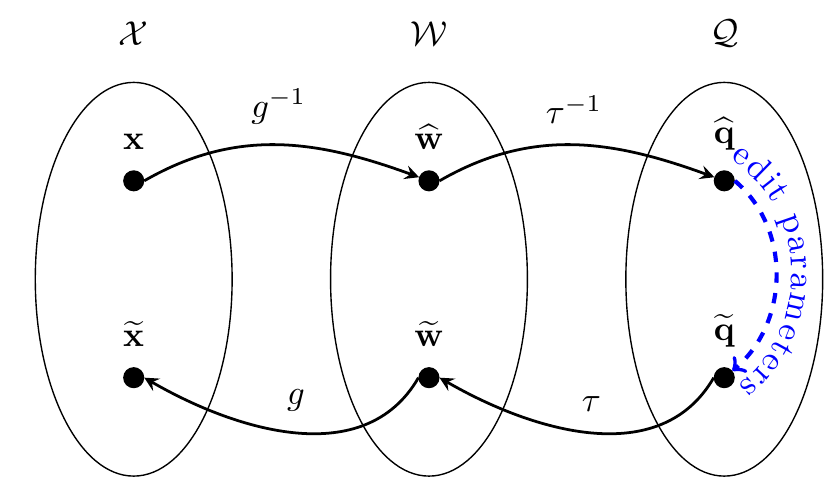}
    \caption{Overview of the different spaces and how the function relate them, c.f.\@ Tab.~\ref{tab:notation}. The blue line indicates a manual change of one of the parameter vectors for transfer of person, expression or rotation.}
    \label{fig:subspaces}
\end{figure}

This section introduces $\tau$GAN, our latent space factorization method for GANs that augments the StyleGAN synthesis network $g$ with a multilinear tensor model.
We do this by embedding a facial expression database into the auxiliary latent space $\mathcal{W}$ of StyleGAN. 
We then order the embedded database into a tensor, which we factorize into semantic subspaces. The resulting parameter space $\mathcal{Q}$ will thus be the Cartesian product of the semantic subspaces $\mathcal{Q} = \mathcal{Q}_\mathrm{P} \times \mathcal{Q}_\mathrm{E}  \times \mathcal{Q}_\mathrm{R}$, where $\mathcal{Q}_\mathrm{P}$ is the person space, $\mathcal{Q}_\mathrm{E}$ the expression space, and $\mathcal{Q}_\mathrm{R}$ is the rotation subspace.
An overview of the different spaces and how the operators relate them are displayed in Fig.~\ref{fig:subspaces} and Tab~\ref{tab:notation}. 

\begin{table}[tb]
\caption{Overview of the notation used in this work.}
\label{tab:notation}
\centering
\begin{tabular}{ll} \toprule 
Symbol & Description     \\ \midrule
$\mathcal{X}$ & Image Space \\
$\mathcal{Z}$ & Latent Space \\
$\mathcal{W}$ & Auxiliary Latent Space \\
$\mathcal{Q}$ & Parameter Space  \\
\bottomrule
\toprule
Operator & Name \\ \midrule 
$f: \mathcal{Z} \to \mathcal{W} $    & Mapping Network  \\
$g:\mathcal{W}\to \mathcal{X} $      & Synthesis Network  \\
$g^{-1}:\mathcal{X}\to \mathcal{W}$  & StyleGAN Embedder  \\
$\tau^{-1}:\mathcal{W}\to \mathcal{Q}$ & Parameter Estimator  \\
$\tau: \mathcal{Q}\to \mathcal{W}$    & Tensor Model   \\    
\bottomrule 
\end{tabular}
\end{table}

\subsection{Tensor Factorization}
The Higher-Order Singular Value Decomposition (HOSVD) is a generalization of the matrix SVD to higher-order tensors \cite{lauth2000,Yano2008,apathy,Multilinearsubspace,tensorface,tensorReview}. 

The starting point for our analysis is a standardized data tensor $\tensor{T}{} \in\mathbb{R}^{N\times P\times E \times R}$, where $N$ refers to the number of elements in the latent vector, $P$ is the number of people, $E$ the number of expressions, and $R$ number of viewpoints or rotations.  
Using the HOSVD $\tensor{T}{}$ can then be factorized as
\begin{align}
\tensor{T}{} \simeq \tensor{C}{} \times_1 \matr{U}_1 \times_2 \matr{U}_2 \times_3 \matr{U}_3 \times_4 \matr{U}_4, 
\label{eq:matrix_hosvd}
\end{align}
where $\times_k$ denotes the $k$-way product, $\tensor{C}{}\in\R^{\tilde{N}\times \tilde{P}\times \tilde{E}\times \tilde{R}}$ is the core tensor, and $\matr{U}_{1}\in\R^{N\times \tilde{N}}$, $\matr{U}_{2}\in\R^{P\times \tilde{P}}$, $\matr{U}_{3}\in\R^{E\times \tilde{E}}$, $\matr{U}_{4}\in\R^{R\times \tilde{R}}$ are matrices with orthonormal columns constructed from the singular vectors of the $k$-mode matrix unfoldings of $T$. In general we have that $\tilde{N}\leq N$, $\tilde{P}\leq P$, $\tilde{E}\leq E$, and $\tilde{R}\leq R$. 

\subsection{Multilinear Tensor Model for GANs} 
The HOSVD \eqref{eq:matrix_hosvd} factorizes the data tensor into a core tensor, and a set of factor matrices $\matr{U}_i$, one for each subspace. 
By selecting appropriate \emph{rows} from $\matr{U}_i$, $i=2,3,4$, one normalized latent vector, i.e. a single mode-1 fiber of $\tensor{T}{}$, can be recovered. 
%To recover exactly one normalized latent vector, i.e. a single mode-1 fiber of $\tensor{T}{}$, we can select the appropriate \emph{rows} of the matrices $\matr{U}_i$ for $i=2,3,4$. 
For example, to recover the latent vector of person $p$ performing expression $e$ with rotation $r$, the $p^\textrm{th}$ row of $\matr{U}_2$, $e^\textrm{th}$ row of $\matr{U}_3$, and $r^\textrm{th}$ row of $\matr{U}_4$ is selected. 
This can be conveniently formulated by a \emph{canonical basis}, where the parameter vectors $\vec{q}'_2\in\R^{P}$, $\vec{q}'_3\in\R^{E}$ and $\vec{q}'_4\in\R^{R}$ pick a weighted linear combination of the rows of the $\matr{U}_i$ matrices. 
Therefore, a given latent code $\vec{y}^\prime$ can be approximated by the model prediction $\widehat{\vec{y}}^\prime$ %which can be written 
as
\begin{align}
\widehat{\vec{y}}^\prime
& = \tensor{C}{}
\times_1 \matr{U}_1 
\times_2 {\vec{q}'_2}^{\T} \matr{U}_2 
\times_3 {\vec{q}'_3}^{\T} \matr{U}_3
\times_4 {\vec{q}'_4}^{\T} \matr{U}_4. 
\label{eq:matrix_model_prime}
\end{align}
This expression can be further simplified by defining $\vec{q}_i^{\T} \equiv {\vec{q}'}_i^{\T} \matr{U}_i$ and analogously $\widehat{\vec{y}}=\matr{U}_1^\T\widehat{\vec{y}}^\prime$. 
Now applying $\times_1\matr{U}_1^\T$ to both sides of \eqref{eq:matrix_model_prime} and recalling that the columns the respective $\matr{U}$ matrices are orthonormal we can write a more compact model representation as
\begin{align}
\widehat{\vec{y}}&= \tensor{C}{}
\times_2 \vec{q}_2^\T
\times_3 \vec{q}_3^\T
\times_4 \vec{q}_4^\T,
\label{eq:matrix_model} 
\end{align}
where the unprimed coordinates refer to the latent code in the eigenspace spanned by the columns of the $\matr{U}_i$ matrices. 
In this formulation, we have 3 individual parameter vectors and use repeated $n$-mode products to relate these to the model prediction.

We can rewrite \eqref{eq:matrix_model} in a more general form to illustrate the mathematical structure of our model. Let us define the $P\times E\times R$, rank-1 parameter tensor $Q=\vec{q}_2^\T\otimes\vec{q}_3^\T\otimes\vec{q}_4^\T$, where $\otimes$ refers to the tensor product. 
Then the components of the rank-1 parameter tensor $Q\in\mathbb{R}^{P\times E \times R}$ is given by $\tensor{Q}{_{\nu}_{\rho}_{\lambda}} = q_{\nu}^{(2)} q_{\rho}^{(3)} q_{\lambda}^{(4)}$ where $q_{\nu}^{(k)}$ refers to the $\nu$th component of the subspace vector $\vec{q}_k\in \mathcal{Q}_k$ for $k=\{2,3,4\}$.

With this definition, we can write \eqref{eq:matrix_model} in a more compact and convenient representation using the Einstein summation convention
\begin{align}
\tensor{\hat{Y}}{^{\mu}} =
\tensor{C}{^\mu^\nu^\rho^\lambda}
\tensor{Q}{_{\nu}_{\rho}_{\lambda}}.
\label{eq:index_model}
\end{align}

This lets us write the latent code, in the auxiliary latent space $\mathcal{W}$, as an application of the multilinear map, defined by the core tensor ${C}$, on the parameter tensor $Q$.

Our entire tensor model $\tau$ can thus be written as the composite map of the core $C$ followed by the change-of-basis transformation defined by $\matr{U}_1:\mathcal{W}\rightarrow \mathcal{W}$, and the inverse standardization operator $\Omega^{-1}:\mathcal{W}\to \mathcal{W}$, where $\Omega^{-1}$ translates and scales a latent vector back to the original scale of $\mathcal{W}$ space according to the mean and variance of the BU-3DFE data.

\subsection{Stacked Style-Separated Model}
\label{sec:extended_model}
In addition to the previously presented model, we propose an alternative approach, where styles are separated instead of vectorizing the latent code. 
That is, we interpret the $S$ styles of $\vec{w}$ as separate vectors of dimension $L$, which is also indicated in Fig.~\ref{fig:stylegan12}. 
To separate the $S$ styles, we propose to order the latent codes into the data tensor $T_\text{style} \in \mathbb{R}^{S\times L\times P \times E \times R}$. 

Then the shape dimension can be addressed separately by defining the style-specific tensors 
\begin{align}
T_s &\in \R^{L\times P \times E \times R}, \quad s = 1,2,\cdots, S. 
\label{eq:tensor18}
\end{align}
We factorize each style-specific tensor $\tensor{T}{_s}$, and define style-specific tensor model $\tau_s$. 
The ensemble of these models is referred to as the \emph{stacked style-separated model} $\tau_S$, which has $S(P+E+R)$ parameters. 
In conclusion, while the prior vectorized model $\tau$, based on $T$, has $P + E + R$ parameters, this formulation  $\tau_S$ has $S(P+E+R)$ parameters since it models the style separately. 

\subsection{Optimization}
\label{subsec:optimization}
Our next aim is to estimate the model parameters by constructing the estimator $\tau^{-1}:\mathcal{W}\to \mathcal{Q}$. 
The estimator is defined as the solution to the optimization problem
\begin{align}
\begin{split}
\min_{Q} \norm{\widehat{\vec{y}}-\vec{y}}_2^2 
\quad &\text{subject to}  \quad \\
\norm{\vec{q}_i}_2^2 \leq c_2 \quad &\text{and}  \quad \\
\norm{\matr{U}_i\vec{q}_i}_1 \leq c_1 \quad &\text{for} \quad i = 2,3,4.
\end{split}
\label{eq:optimization_problem}
\end{align}
The form of \eqref{eq:optimization_problem} is inspired by \cite{apathy, grasshof2020}, and enforces constraints on the model parameters to retrieve a stable representation of new latent vectors by linear combinations within the training data. We regularize the model using Ridge and Lasso regression. 
%and regularize the model by Elastic Net \cite{elasticnetZou2005}. 
Then the Lagrangian for the constrained problem \eqref{eq:optimization_problem} can be written as
\begin{align}
\begin{split}
\mathcal{L}(Q,\lambda_1,\lambda_2) 
&= \norm{\widehat{\vec{y}}-\vec{y}}_2^2 \\ 
&+\sum_{k=2}^4 \lambda_{2,k}\norm{\vec{q}_k}_2^2 
+ \lambda_{1,k}\onenorm{\vec{q}^\prime_k}
\label{eq:lagrangian}
\end{split}
\end{align}
where $\lambda_{1,k}, \lambda_{2,k}\geq0$ refer to regularization parameters, i.e. Lasso and Ridge. Note that there is no prime on the Ridge term since $\norm{\vec{q}^\prime_i}_2^2 = (\matr{U}_i^\T\vec{q}_i)^\T(\matr{U}_i^\T\vec{q}_i) = \norm{\vec{q}_i}_2^2$ since $\matr{U}_i^\T \matr{U}_i = \matr{I}$. 
We will now continue to present a strategy for solving the constrained optimization problem in \eqref{eq:optimization_problem} by Alternating Least Squares. 

As in \cite{apathy,grasshof2020} the minimization can be solved by first rewriting \eqref{eq:matrix_model} as a matrix-vector multiplication separately for each of the three model parameter vectors as
\begin{align}
\hat{\vec{y}} = \matr{A}^{(k)}\vec{q}_{k}, ~k=2,3,4,
\label{eq:als}
\end{align}
where the matrices $\matr{A}^{(k)}$ are given by
\begin{align}
\matr{A}^{(2)} &= C \times_3 \vec{q}_3^\T\times_4 \vec{q}_4^\T, \\
\matr{A}^{(3)} &= C \times_2 \vec{q}_2^\T\times_4 \vec{q}_4^\T, \\
\matr{A}^{(4)} &= C \times_2 \vec{q}_2^\T\times_3 \vec{q}_3^\T.
\end{align}
Therefore, an unknown latent vector $\vec{y}$ can be estimated by alternating between the systems \eqref{eq:als}, while updating the matrices $\matr{A}^{(k)}$ in each step.

%%%%%%%%%%%%%%%%%%%%%%%%%%%%%%%%%%%%%%%
%%%%% PART3 Experiments,
%  Subspace, Validation, Parameter Transfer
%  IFGAN comparison 
%%%%%%%%%%%%%%%%%%%%%%%%%%%%%%%%%%%%%%%

\section{Experiments}\label{sec:results}
In the following, we give some additional details for the BU-3DFE database and continue to report on our experimental results.

\subsection{Facial Expression Database} \label{sec:data}
As mentioned in the introduction, we use the BU-3DFE database \cite{bu3dfe}. 
The database contains 3D face scans and images of 100 persons (56 female and 44  male), with varying ages (18-70 years) and diverse ethnic/racial ancestries. 
Each subject was asked to perform the six basic emotions: anger, disgust, happiness, fear, sadness, and surprise, each with four levels of intensity. 
Additionally, for each participant, the neutral face was recorded. 
Hence, for each person, there are a total of 25 facial expressions recorded from two pose directions, left and right, resulting in 5000 face images.

\subsection{Data Prepossessing}
As a pre-processing step, we embedded each face image from the BU-3DFE database, into the latent space of StyleGAN, as described in Sec.~\ref{subsec:generator_inversion}. 
We then collected the resulting latent vectors into the 4-way data tensor $T_0\in\mathbb{R}^{N\times P\times E\times R}$. 
We then calculated the mode-$1$ unfolding $\matr{T}_0^{(1)}\in\mathbb{R}^{N\times PER}$ of $T_0$ containing all the $PER$ latent vectors. We then standardized this matrix to zero mean and unit variance for each latent variable and then finally folded this standardized matrix into a $N\times P\times E\times R$ dimensional tensor $\tensor{T}{}$ which we used for all subsequent experiments.

% The standardized mode-1 unfolding thus is
% \begin{align}
% \matr{T}^{(1)} &= \matr{\Lambda}^{-1/2}\left(\matr{T}_0^{(1)}-\matr{M}\right)
% \label{eq:standarize}
% \end{align}
% where $\matr{\Lambda}\in\R^{N\times N}$ is a diagonal matrix containing the $N$ row sample variances of $\matr{T}_0^{(1)}$ on the diagonal, and $\matr{M}=\frac{1}{n}\matr{T}_0^{(1)}\vec{1}_{n}\vec{1}_{n}^\T$, $n=PER$ is the matrix which contains the mean latent code in each column repeated.
%\subsection{Decomposition}

\subsection{Subspace Analysis} \label{sec:subspace}
\begin{figure}[tb]%[h!]
    \centering
    \includegraphics[width=\linewidth]{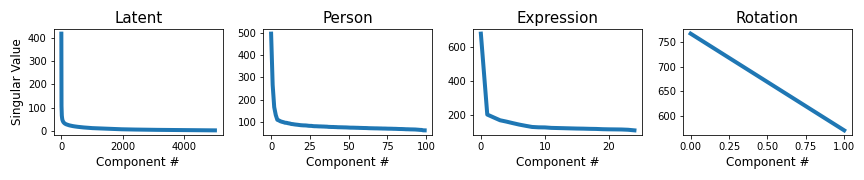} 
\caption{Validation of decomposition results. Energy of singular values for each mode of $\tensor{T}{}$.} 
\label{fig:energy}
\end{figure}

The standardized tensor $\tensor{T}{}$ was factorized by the HOSVD, as described in \eqref{eq:matrix_hosvd}, yielding the four subspaces spanned by the columns of $\matr{U}_k$, $k=1,\ldots,4$. The distribution of the energy of the subspaces is shown in Fig.~\ref{fig:energy}, which illustrates the compactness of the subspaces. 

\begin{figure}[tb]
    \centering
    \includegraphics[width=0.6\linewidth]{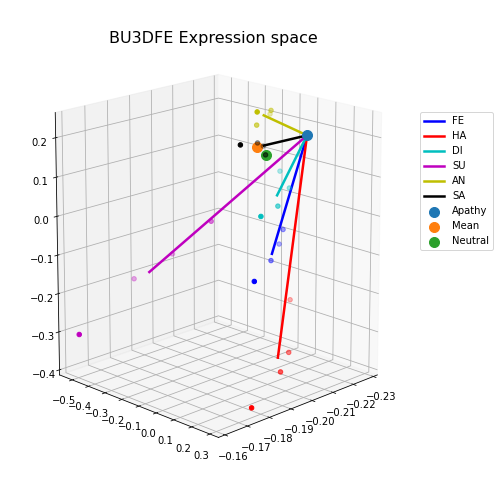} 
    % \begin{subfigure}{.49\linewidth}        
    %     \includegraphics[width=\linewidth]{fig/apathy/apathy2d.png}  
    %     \caption{2D projection}
    %     \label{subfig:subspaces2d}
    % \end{subfigure}
    % \begin{subfigure}{.49\linewidth}
    %     \includegraphics[width=\linewidth]{fig/apathy/apathy3d.png}  
    %     \caption{3D projection}
    %     \label{subfig:subspaces3d}
    % \end{subfigure}
\caption{
    Projection of the expression subspace,  defined by $\matr{U}_3$, onto 3 dimensions.
% Projection of the expression subspace,  defined by $\matr{U}_3$, onto (\subref{subfig:subspaces2d}) 2,  
% and (\subref{subfig:subspaces3d}) 3 dimensions.
} 
\label{fig:exprsubspaces}
\end{figure}

\begin{figure}[tb]
\centering
\begin{subfigure}{.3\linewidth}
    \includegraphics[width=\linewidth]{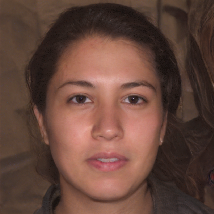} \caption{Mean}
    \label{subfig:mean}
\end{subfigure}
\begin{subfigure}{.3\linewidth}
    \includegraphics[width=\linewidth]{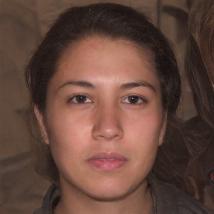}  
    \caption{Neutral}
    \label{subfig:neutral}
\end{subfigure}
\begin{subfigure}{.3\linewidth}
    \includegraphics[width=\linewidth]{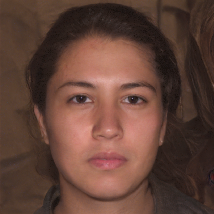}  
    \caption{Apathy}
    \label{subfig:apathy}
\end{subfigure}\\
% \hfill
% \centering

\begin{subfigure}{.22\linewidth}
    \includegraphics[width=\linewidth]{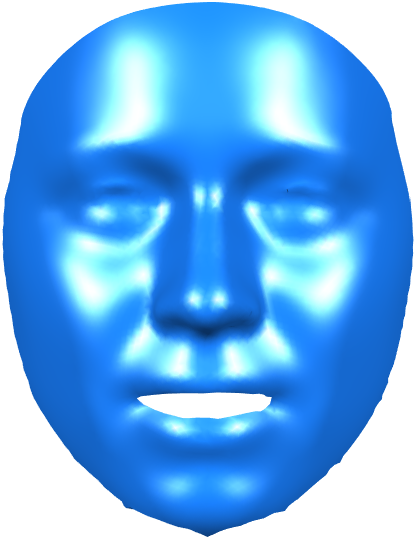}
    \caption{Mean}
    \label{subfig:mesh_mean}
\end{subfigure}
% \hfill
\hspace{15pt}
\begin{subfigure}{.22\linewidth}
    \includegraphics[width=\linewidth]{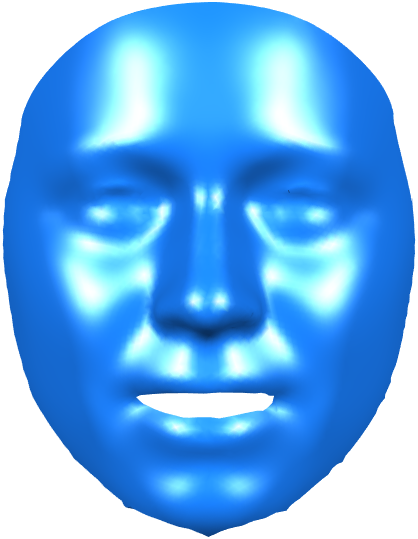}
    \caption{Neutral}
    \label{subfig:mesh_neutral}
\end{subfigure}
% \hfill
\hspace{15pt}
\begin{subfigure}{.22\linewidth}
    \includegraphics[width=\linewidth]{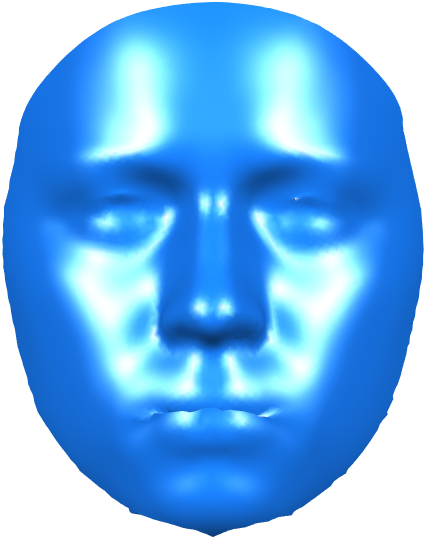}
    \caption{Apathy}
    \label{subfig:mesh_apathy}
\end{subfigure}
\hfill 
\caption{Synthesized faces for (\subref{subfig:mean}) the mean face, (\subref{subfig:neutral}) neutral face, and (\subref{subfig:apathy}) apathetic face. 
Accordingly, (\subref{subfig:mesh_mean}), (\subref{subfig:mesh_neutral}), (\subref{subfig:mesh_apathy}) show the 3D faces synthesized by the method in \cite{apathy}.}
\label{fig:apathy_vs_neutral} 
\end{figure}

In Fig.~\ref{fig:exprsubspaces} we show a visualization of the expression subspace.
As an initial step, we truncated the expression subspace from 25 dimensions to 3D.  
It can be seen that for each emotion, the variation in expression strength forms linear trajectories in expression space. 
These trajectories are star-shaped and meet at an origin of expression which is shared by all emotion trajectories. 
This is neither the neutral nor the mean face, but the ``apathetic'' face, found in \cite{apathy,grasshof2020}, see Fig.~\ref{fig:apathy_vs_neutral}(\subref{subfig:mean})-(\subref{subfig:apathy}). 
In this case, the apathetic face in Fig.~\ref{fig:apathy_vs_neutral}(\subref{subfig:apathy}) is closer to the mean face than in \cite{apathy,grasshof2020}, displayed in Fig.~\ref{fig:apathy_vs_neutral}(\subref{subfig:mesh_apathy}) for comparison.

\subsection{Vectorized vs. Stacked Style-Separated Model}\label{sec:extmodel}

\begin{figure}[tb]
\centering
%%%%%%%%%%%%%%%% REFERENCE    
\begin{subfigure}{.30\linewidth}
\includegraphics[width=\linewidth]{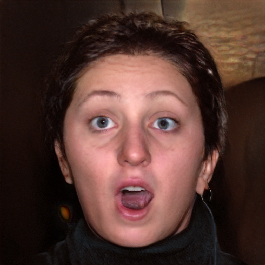}\\
\includegraphics[width=\linewidth]{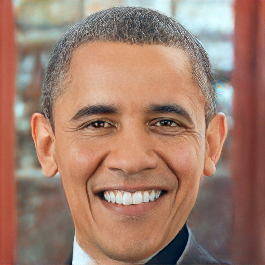}\\
\includegraphics[width=\linewidth]{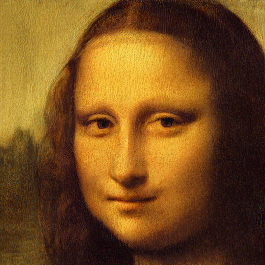}     
\caption{Reference}
\label{subfig:reference}
\end{subfigure}
%%%%%%%%% Vectorized 
\begin{subfigure}{.30\linewidth}
\includegraphics[width=\linewidth]{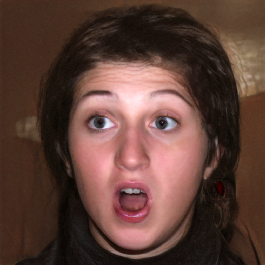}\\
\includegraphics[width=\linewidth]{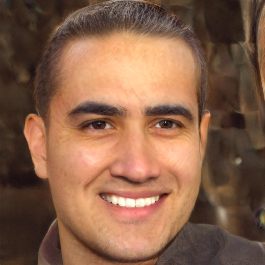}\\
\includegraphics[width=\linewidth]{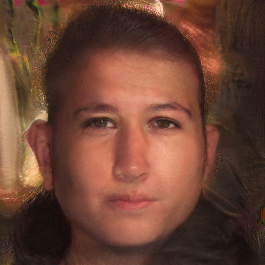}
\caption{Vectorized}
\label{subfig:vectorised}
\end{subfigure}
%%%%%%%%%%%%%%%%
\begin{subfigure}{.30\linewidth}
\includegraphics[width=\linewidth]{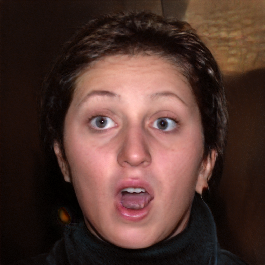}\\
\includegraphics[width=\linewidth]{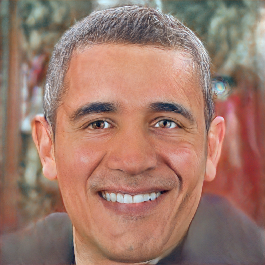}\\
\includegraphics[width=\linewidth]{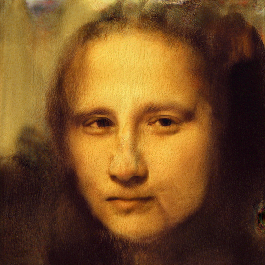}
\caption{Style-Separated}     
\label{subfig:extended}
\end{subfigure}
\caption{Reconstructions: (\subref{subfig:reference}) Ground truth images, and the  results from either (\subref{subfig:vectorised}) the vectorized model, and (\subref{subfig:extended}) the style-separated model. The top row shows an example from the BU-3DFE database, while the 2nd and 3rd rows illustrate reconstruction of novel images which are not part of BU-3DFE. 
}
\label{fig:fig7}
\end{figure}

In Sec.~\ref{sec:tensormodel} we proposed to build two different versions of tensor models. 
(1) The \emph{vectorized model} flattens each latent code of one image and then orders them into the tensor $T\in\R^{N\times P\times E\times R}$, and 
(2) the \emph{stacked style-separated model} $T_\text{style}\in\R^{S\times L\times P\times E\times R}$ which considers the $S=18$ styles of StyleGAN separately.
We estimated the parameters for the two models, using the ALS procedure \eqref{eq:als}.
The results are illustrated in Fig.~\ref{fig:fig7}. 
It can be seen, that the ground truth (Fig.~\ref{fig:fig7}\subref{subfig:reference}), is visually closer to the stacked style-separated model (Fig.~\ref{fig:fig7}\subref{subfig:extended}) than the vectorized model (Fig.~\ref{fig:fig7}\subref{subfig:vectorised})
for test images from the BU-3DFE data set (top row), as well as for arbitrary images (2nd and 3rd row). 
We conclude that the proposed adaptation by the separate styles improves performance. 

\subsection{Validation of Regularization Parameters}
\label{sec:validation}
The optimization problem defined in \eqref{eq:lagrangian} contains six regularization parameters $\lambda_{1,k}$ and $\lambda_{2,k}$, $k=2,3,4$, two for each of the three parameter vectors, which must be manually set. 
In the following experiment we investigated how the hyperparameters influenced the quality of the results, and assume that they are the same for the three parameters, hence $\lambda_1=\lambda_{1,k}$, and $\lambda_2=\lambda_{2,k}$.
Here we used the vectorized model on the basis of the standardized latent codes in \eqref{eq:matrix_model}. 
Initially, we divided the data into a training, validation, and test set by a randomized 90--5--5 split over the $P=100$ person identities.
The validation set thus had a total of $5ER=250$ samples. 
We estimated the tensor model based on the training set. 
For each latent vector in the validation set we then estimated the subspace parameters $\vec{q}_i$ by ALS using \eqref{eq:als}.

We evaluated three kinds of errors for the validation set: the approximation error, and the expression and rotation transfer errors.
The approximation error between the ground truth $\vec{y}$ and estimated latent code $\widehat{\vec{y}}_i$ is defined as $\epsilon_\text{approx}=\norm{\widehat{\vec{y}}-\vec{y}}^2_2$. 
The transfer errors result from exchanging estimated parameters $\widehat{\vec{q}}_k$ by known values $\widetilde{\vec{q}}_k$. 
Hence using $\widetilde{\vec{y}}_\text{expr} \equiv \tau(\widehat{\vec{q}}_{\text{person}} \otimes \widetilde{\vec{q}}_{\text{expr}}\otimes \widehat{\vec{q}}_{\text{rot}})$ gives rise to an expression transfer error which we define as $\epsilon_\text{expr}=\norm{\widetilde{\vec{y}}_\text{expr}-\vec{y}}^2_2$. Analogously, the rotation transfer error is defined as the error arising from only changing the parameters associated with the rotation subspace according to $\epsilon_\text{rot} = \norm{\widetilde{\vec{y}}_\text{rot}-\vec{y}}_2^2$. 
The three error metrics $\epsilon_\text{approx}$, $\epsilon_\text{expr}$, and $\epsilon_\text{rot}$ were then calculated for each sample, with varying hyperparameter values $\lambda_1$ and $\lambda_2$. 
In this experiment, we investigate Lasso and Ridge regression independently, i.e., we set $\lambda_1 = 0$ while varying $\lambda_2$, and vice versa. We restrict ourselves to only consider cases where the regularization strength is equal for all subspaces.

\begin{figure}[tb]
    \centering
    \begin{subfigure}{\linewidth} % left, bottom, right
    \includegraphics[width=0.99\linewidth]{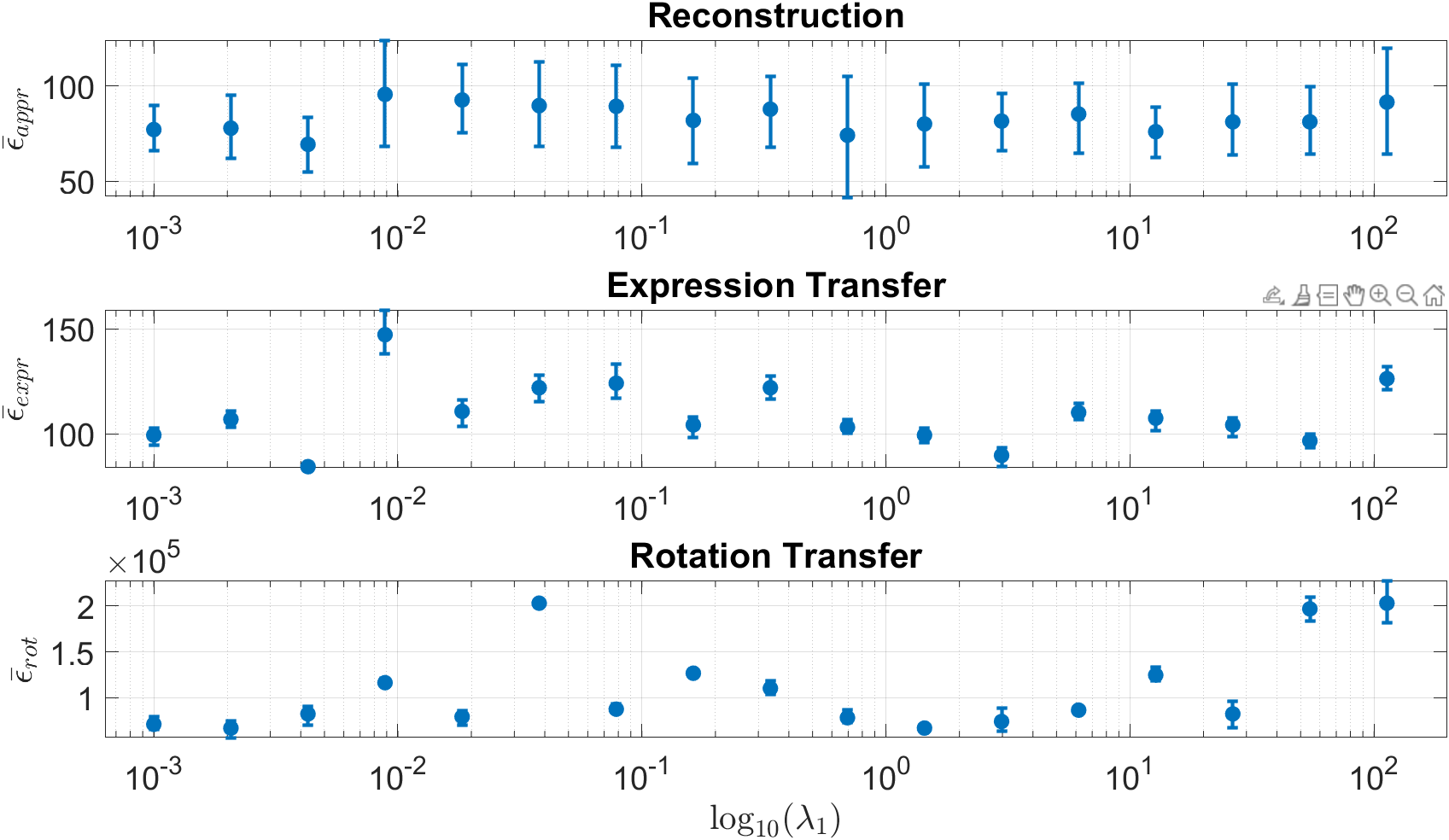}
    \caption{Lasso (L1 penalty)}
    \label{subfig:l1penalty}
    \end{subfigure}

    \begin{subfigure}{0.99\linewidth}
    \includegraphics[width=0.99\linewidth]{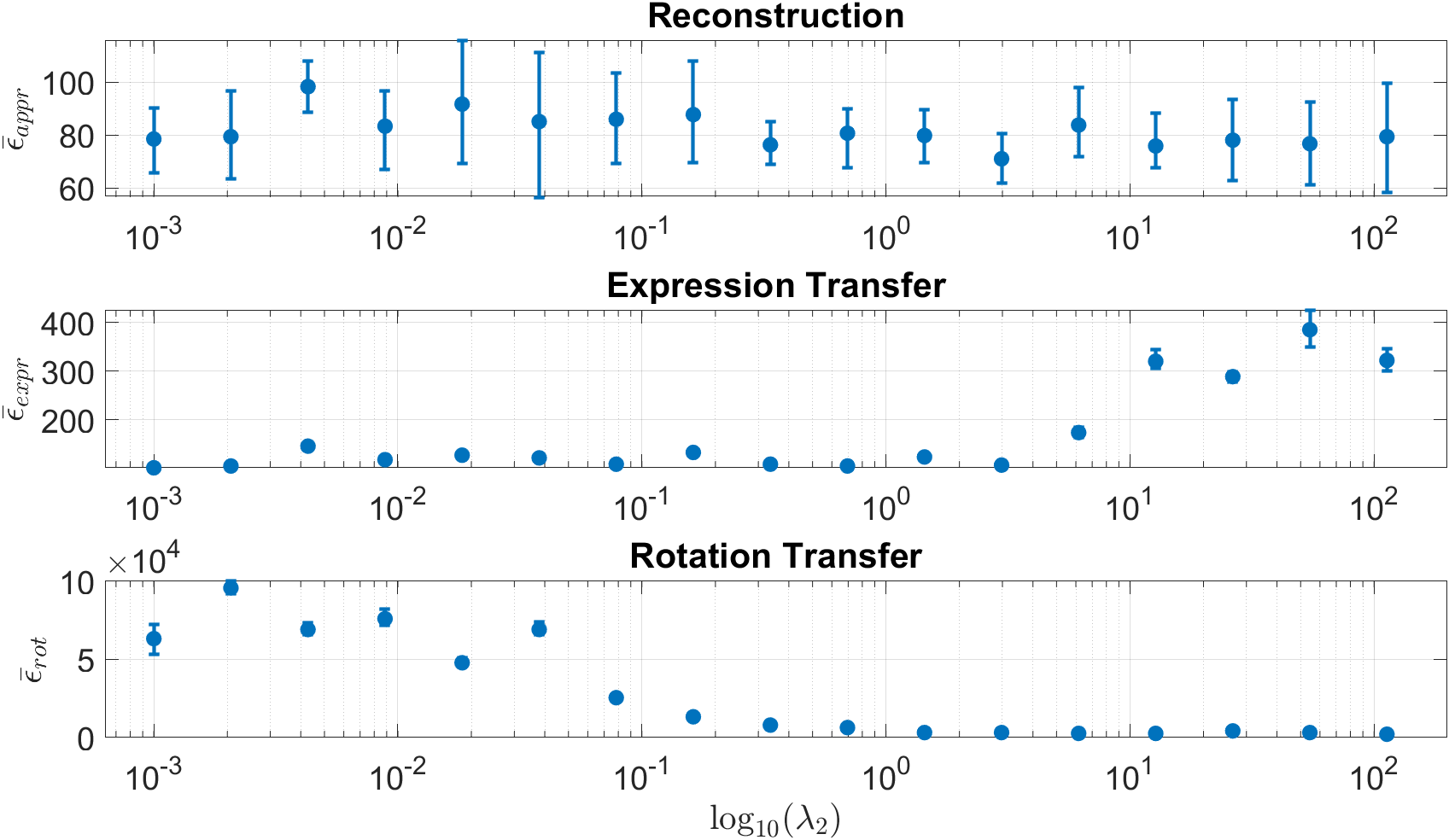} 
    \caption{Ridge (L2 penalty)}
    \label{subfig:l2penalty}
    \end{subfigure}
\caption{Influence of the hyper parameters, $\lambda_1$ and $\lambda_2$ steering the (\subref{subfig:l1penalty}) Lasso and  (\subref{subfig:l2penalty}) Ridge constraints, on (from top to bottom row) the approximation error, expression transfer error, and rotation transfer error.}
\label{fig:hyper_param_errors}
\end{figure}

The results are illustrated in Fig.~\ref{fig:hyper_param_errors}. 
In general, it can be seen that the approximation error is more stable than the other two errors.
Fig.~\ref{fig:hyper_param_errors}\subref{subfig:l1penalty} suggests that high values of $\lambda_1$ should be chosen for rotation transfer, while for expression transfer $\lambda_1\approx 1$ seems to be a reasonable choice. 
Fig.~\ref{fig:hyper_param_errors}\subref{subfig:l2penalty} reveals that for $\lambda_2 \approx 1$ all error metrics are small, and hence this interval is a good choice.  

\subsection{Regularization and Parameter Transfer} \label{sec:approx}
\begin{figure}[tb]  
\centering
\begin{subfigure}{.3\linewidth}
\centering
\includegraphics[width=\linewidth]{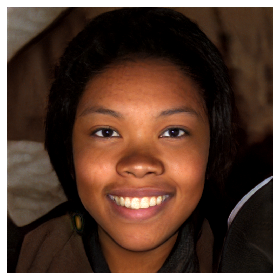}  
\caption{Ground Truth}\label{fig9:true}
\end{subfigure}
\begin{subfigure}{.3\linewidth}
\centering
\includegraphics[width=\linewidth]{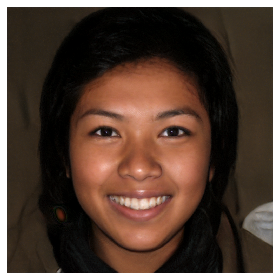}  
\caption{Non-regularized}\label{fig9:ols}
\end{subfigure}
\begin{subfigure}{.3\linewidth}
\centering
\includegraphics[width=\linewidth]{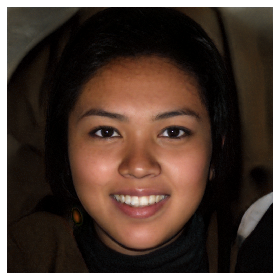}  
\caption{Regularized}\label{fig9:reg}
\end{subfigure}

\begin{subfigure}{.49\linewidth}
\centering
%\rotatebox[]{90}{OLS}
\includegraphics[width=\linewidth]{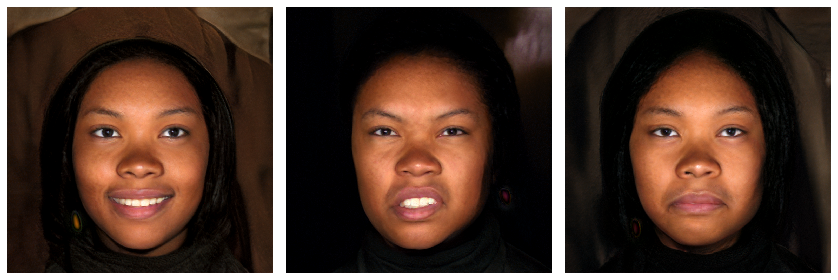}  
\includegraphics[width=\linewidth]{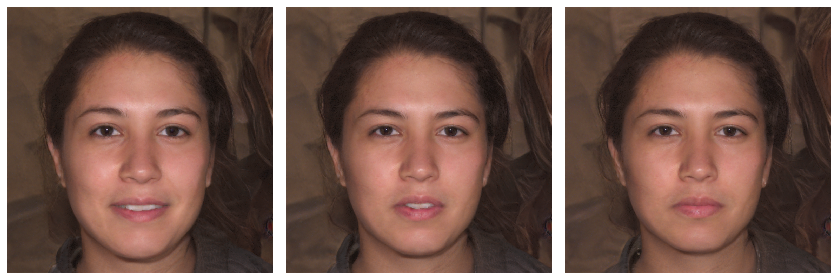}  
\includegraphics[width=\linewidth]{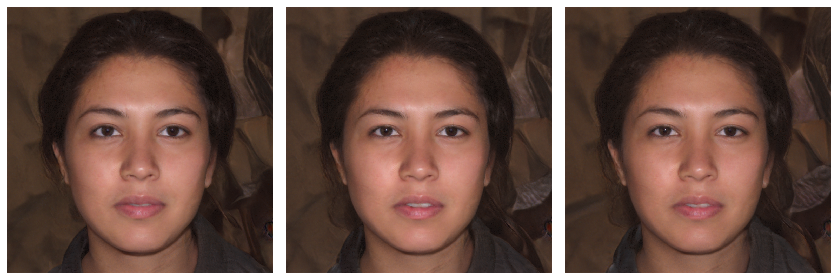}  
\caption{Expression Transfer}\label{fig9:exp}
\end{subfigure}
\begin{subfigure}{.49\linewidth}
\centering
\includegraphics[width=\linewidth]{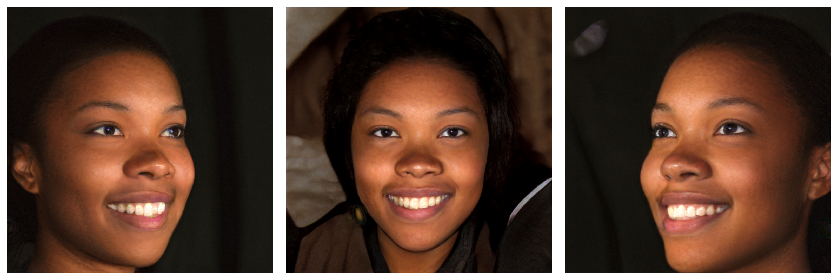}
\includegraphics[width=\linewidth]{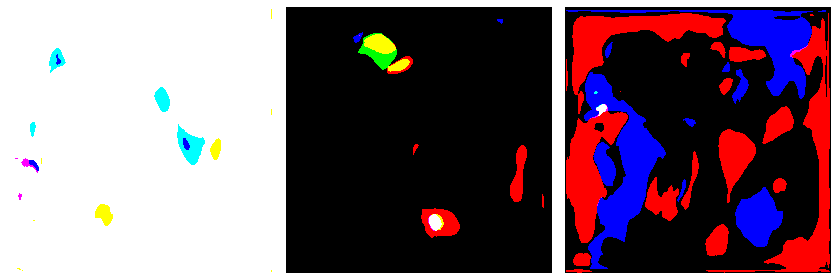}  
\includegraphics[width=\linewidth]{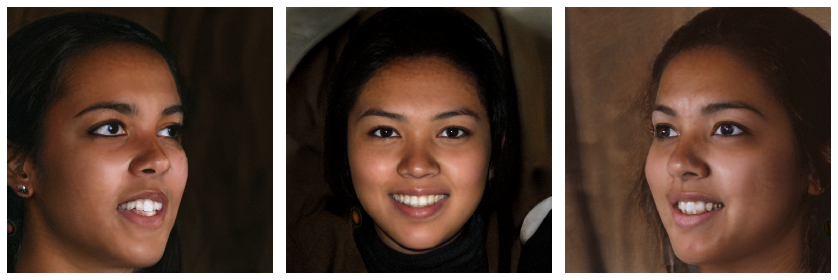}  
\caption{Rotation Transfer}\label{fig9:rot}
\end{subfigure}
% Figure  \ref{fig9:true}, \ref{fig9:ols} and \ref{fig9:reg}.  
\caption{Reconstruction and regularization results. (\subref{fig9:true}) Ground truth (\subref{fig9:ols}) approximation by the non-regularized model, and (\subref{fig9:reg}) the regularized  model.
(\subref{fig9:exp},\subref{fig9:rot}) Results from rotation and expression transfer containing  ground truth (top row), the non-regularized solutions (middle row), and the regularized solution  (bottom row).}
\label{fig:fig9}
\end{figure}

We used the regularization parameters above to perform expression and rotation transfer on samples from the test set.
We then synthesized images from the estimated parameters by applying the composite transformation $\widehat{\vec{x}}= g(\tau(\widehat{Q}))$ to the estimated subspace parameters $\widehat{Q}$. 
Additionally, we performed expression and rotation transfer by replacing one of the three estimated parameter vectors by known values, as described before.
We did this for the regularized model ($\lambda_1>0,~\lambda_2>0$) and the non-regularized model ($\lambda_1=\lambda_2=0$). 
Fig.~\ref{fig:fig9} shows how well the ground truth, in $\mathcal{W}$ space,  (Fig.~\ref{fig:fig9}\subref{fig9:true}) can be approximated by the non-regularized solution (Fig.~\ref{fig:fig9}\subref{fig9:ols}) and the regularized solution (Fig.~\ref{fig:fig9}\subref{fig9:reg}). 
It seems that the non-regularized solution matched the ground truth slightly better with respect approximation expression transfer. 
However, for rotation transfer (Fig.~\ref{fig:fig9}\subref{fig9:rot}) the regularized solution clearly outperformed the non-regularized solution.
Because in the non-regularized solution the resulting image is not recognizable as a face anymore at all, while the regularized solution is not deformed and the rotation of the depicted faces conform to ground truth.
This experiment thus showed that adding a small L2 regularization term yields stable rotation transfer. 

\subsection{Quantitative Comparison}

\begin{figure}[tb]  
\centering
\begin{subfigure}{.49\linewidth}
\includegraphics[width=\linewidth]{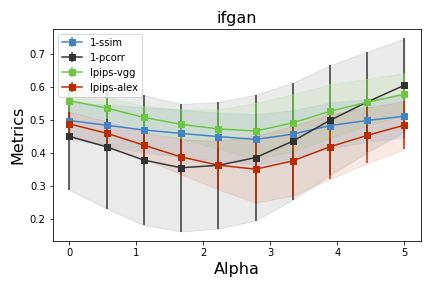}  
\end{subfigure}
\begin{subfigure}{.49\linewidth}
\includegraphics[width=\linewidth]{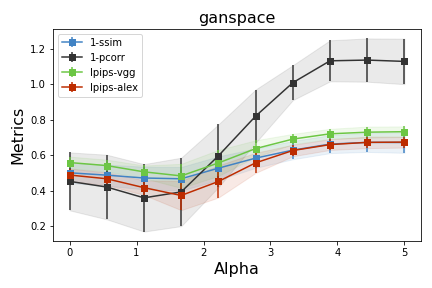}  
\end{subfigure}
\caption{
To find the optimal interpolation strength $\alpha$ for rotation transfer for  InterFaceGAN \cite{Shen2020interfacegan} and GANSpace \cite{Harkonen2020GANSpace} we compare the images generated by shifting the latent code corresponding to an image from the one rotation towards the other and compare the result with the ground truth. 
% Results of varying values of $\alpha$ for face editing with InterFaceGAN \cite{Shen2020interfacegan}.
} 
\label{fig:ifgan-hyper}
\end{figure}

Finally, we compare $\tau$GAN to InterFaceGAN \cite{Shen2020interfacegan} and GANSpace \cite{Harkonen2020GANSpace} for the application of semantic face editing by using rotation transfer as one example.

Since the BU-3DFE database \cite{bu3dfe}, see Sec.~\ref{sec:data}, contains 5000 faces images, 2500 from the left and from the 2500 right; we chose one of the two views as the reference image, and then used InterFaceGAN, GANSpace and  $\tau$GAN to estimate a reconstruction of the image from the complementary rotation.
The resulting image was then compared to the Ground Truth (GT) by
1) Pearson correlation coefficient (pcorr), 2) Structural Similarity Index Measure (SSIM) \cite{Wang2002SSIM}, and 3) Learned Perceptual Image Patch Similarity (LPIPS) \cite{Zhang2018unreasonable}. 
For the LPIPS measure, we employed two versions: one based on VGG \cite{vgg16}, referred to as lpips-vgg, and the other, lpips-alex, on AlexNet \cite{Krizhevsky2012Alexnet}.

% IFGAN 
In InterFaceGAN \cite{Shen2020interfacegan} the authors find semantic directions of StyleGAN by fitting SVMs to single semantic attributes using an annotated data set. 
Using these directions, semantic editing can be performed by interpolating in the direction  $\vec{n}\in\mathbb{R}^N$ defined by the SVM hyper-plane normal vector for a given latent code $\vec{w}\in\mathcal{W}$, as 
\begin{align}
    \vec{w}_\text{edit} = \vec{w} + \alpha \vec{n},
    \label{eq:ifgan}
\end{align}
where $\alpha$ is the strength of the shift in semantic direction associated with $\vec{n}$. 
To perform rotation transfer, we chose the pose direction for the StyleGAN1 model trained on FFHQ provided by \cite{Shen2020interfacegan} as $\vec{n}$. 

%%% GANSpace
GANSpace finds semantic directions in an unsupervised fashion using PCA. 
The semantic meaning of the found principal components needs to be assigned by a one-time manual labeling. 
In the paper the authors report that the 10th principal component applied only to the first 7 layers produces a shift in rotation for the pretrained StyleGAN1 network. 
Using this definition, and the rotation direction, we can perform semantic edits with GANSpace in a similar way as in eq.~\ref{eq:ifgan}.

To determine the optimal interpolation strength $\alpha$ for both methods, we design an experiment where we perform rotation transfer with varying values for $\alpha$. 
From the latent code representing an image of one rotation, we edit the latent code towards the complementary rotation resulting in a latent vector $\vec{w}_\text{edit}$ which is then used to synthesize an edited image. 
We then compare the edited image to the ground truth using the four metrics mentioned above.
For each value of $\alpha$ we average the metrics and pick the minimum. 
The results are presented in Fig.~\ref{fig:ifgan-hyper}, where it can be seen that the best performance for InterFaceGAN is reached at $\alpha = 2.77$, and for GANSpace at $\alpha = 1.66$, respectively. 
%We found that $\alpha = 2.77$ and $\alpha = 1.66$ gives the best performance for InterFaceGAN and GANSpace respectively. 
These values are used for the quantitative comparison presented in Fig.~\ref{fig:ifgan-rot-compare}. 
%in the following experiments.

\begin{figure}[tb]
\centering
\begin{subfigure}{0.195\linewidth}
\includegraphics[width=\linewidth]{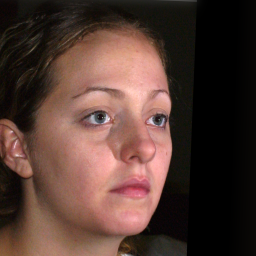}  
\end{subfigure}  
\hspace{143pt}
\begin{subfigure}{0.195\linewidth}
\includegraphics[width=\linewidth]{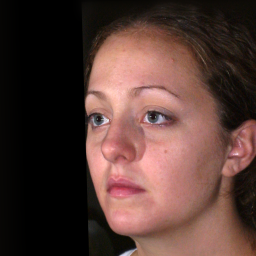}
\end{subfigure}

\begin{subfigure}{\linewidth}
\includegraphics[width=\linewidth]{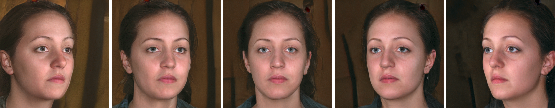}
\caption{Direct Interpolation}
\end{subfigure}
\begin{subfigure}{\linewidth}
\includegraphics[width=\linewidth]{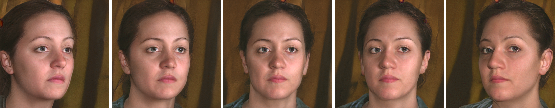}  
\caption{InterFaceGAN}
\end{subfigure}  
\begin{subfigure}{\linewidth}
\includegraphics[width=\linewidth]{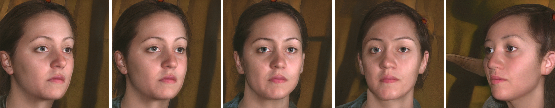}
\caption{GANSpace}
\end{subfigure}  
\begin{subfigure}{\linewidth}
\includegraphics[width=\linewidth]{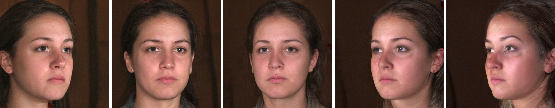}  
\caption{$\tau$GAN (Ours)}
\end{subfigure}
\caption{
%Visual comparison between rotations produced by varying methods.
Comparison of rotation transfer among varying methods. 
%InterFaceGAN and our proposed $\tau$GAN model.
The ground truth images in pixel space are shown in the top row in the outermost columns. We use the latent code corresponding to the left hand rotation (top left) and try to recover the right hand rotation (top right). 
The provided images have been created by: 
(a) direct interpolation, (b) InterFaceGAN, (c) GANSpace, and (d) our proposed $\tau$GAN.
%Here we make a visual comparison between rotations produced by direct interpolation (second row), InterFaceGAN, GANSpace and our proposed $\tau$GAN.
}
\label{fig:ifgan-visual}
\end{figure}

\begin{figure}[tb]
    \centering
    \includegraphics[width=\linewidth]{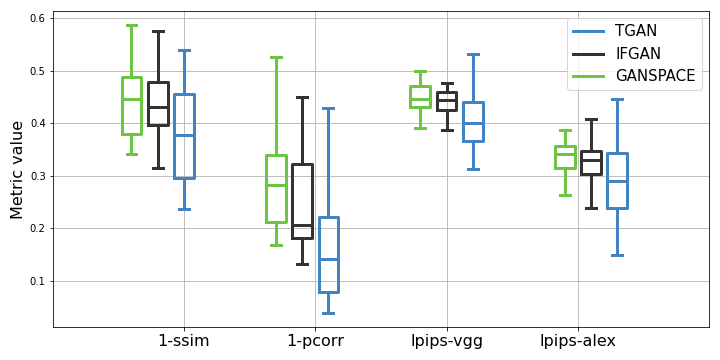}
    \caption{Quantitative comparison of rotation transfer performed by varying methods. 
    We start with images from the left rotation and shift the latent codes towards the right rotations using $\tau$GAN, InterFaceGAN, and GANSpace. 
    The edited images are then compared to the GT based on the previously used adapted metrics, redefined to be the lower the better. 
%    We then compare the edited image to the known ground truth using the four metrics described in the main text. 
%    The metrics are defined as the lower the better. 
    We observe that the edited images produced by $\tau$GAN are more similar to the GT %ground truth 
    across all four metrics. 
    }
    \label{fig:ifgan-rot-compare}
\end{figure}

%%% tGAN
To perform rotation transfer with $\tau$GAN model, we first estimated the model parameter vectors $\widehat{\vec{q}}_k$, $k=2,3,4$ for a given input image as described in Sec.~\ref{subsec:optimization}.
Then we used the rotation subspace defined by $\matr{U}_4$ in \eqref{eq:matrix_hosvd}. 
For $\tau$GAN we take the subspace direction $\vec{m} = \vec{u}_2^{(4)} - \vec{u}_1^{(4)}\in \mathcal{Q}_\mathrm{R}$, where $\vec{u}_1^{(4)}$, $\vec{u}_2^{(4)}$ are the first and second row of $\matr{U}_4$, respectively. 
The rotation parameter was then changed as 
\begin{align}
\widetilde{\vec{q}}_4 = \widehat{\vec{q}}_4+\gamma\vec{m},
\end{align}
which then yields the edited latent code
\begin{align}
\vec{w}_\text{$\tau$,edit} = \tau(\widehat{\vec{q}}_2 \otimes \widehat{\vec{q}}_3\otimes \widetilde{\vec{q}}_4).
\end{align}
%  
%%%%%%%%%%%%%%%%%%%%%%%%%%%%%%%%%%
% $\vec{w}_\text{IF,edit}$ and $\vec{w}_\text{$\tau$,edit}$
%%% Qualitative experiment
Fig.~\ref{fig:ifgan-visual} shows synthesized images produced by InterFaceGAN, GANSpace and $\tau$GAN, respectively. 
These are compared against the reconstructions generated by latent codes interpolated directly in $\mathcal{W}$ space by $\vec{w} = \beta \vec{w}_\text{left} + (1-\beta)\vec{w}_\text{right}$ where $\vec{w}_\text{left}$ and $\vec{w}_\text{right}$ refer to the left and right rotation, respectively. 
The results show that $\tau$GAN provides an alternative way for generating rotation in the StyleGAN latent space. 
Compared to InterFaceGAN, our model seems to create rotations which better preserve features like skin tone and gaze direction, and compared to GANSpace the face shape seems better preserved. 
However, for all methods we note that the identity of the person slightly changes in this example. % when performing rotations. 

%%% Quantitative experiment
%%%%%% NEW START
%To evaluate the rotation transfer, we apply the previously introduced three methods to the task: InterFaceGAN, GANSpace, and our proposed $\tau$GAN. 
Additionally, %to the qualitative results provided in Fig.~\ref{fig:ifgan-visual}, 
we objectively compare the quality of rotation transfer resulting from different methods as follows.
We apply the previously introduced three methods: InterFaceGAN, GANSpace, and our proposed $\tau$GAN, to shift the rotation of the 125 left-oriented images in the validation set towards the right orientation. 
We then compare the edited images to the known ground truth using the same four metrics introduced at the beginning of this section. 
The results in Fig.~\ref{fig:ifgan-rot-compare} show that $\tau$GAN has the lowest median value for all metrics when compared with %the competing
InterFaceGAN and GANSpace. % hence outperforms them.
%%%%%% NEW END
%%%%%% OLD START
%Additionally,  we  make  a  quantitative  comparison  of the rotations produced by the three methods which are presented in Fig. ~\ref{fig:ifgan-rot-compare}.  
%Here we use the respective methods to shift the rotation of the 125 left oriented images in the validation set towards the right orientation. We then compare the edited images to the known ground truth using the same 4 metrics. It can be seen that, for each metric $\tau$GAN has lower median metric values when compared with InterFaceGAN and GANSpace. 
%%%%%% OLD END

% It can be seen that the rotations produced by our approach are closer to the BU-3DFE ground truth images. 
% Also for each of the four metrics,  . 
% Quantitatively we compare the rotations 
% The rotations produced by the three methods were compared to ground truth latent codes, as follows. 
% The quantitative comparison results are 

\section{Conclusions}\label{sec:conclusion}

%Summary 
In this work, we proposed $\tau$GAN, a tensor-based model for the auxiliary latent space of the StyleGAN. It is constructed by first embedding the images of the BU-3DFE database into the latent space of StyleGAN.
The latent codes were stored into a tensor which is then factorized into semantically meaningful subspaces by HOSVD. 
This construction ensured that the semantic directions were directly interpretable in contrast to unsupervised methods, where this is not always the case. 

% Compare to apathy 
We were able to generalize previous results \cite{apathy} of face analysis by showing that the expression subspace has the structure where the expression trajectories meet in a specific \emph{apathetic} expression, which is different from the mean or neutral face. 
% Conclusion
We evaluated our approach quantitatively and qualitatively, and compared different versions of the proposed tensor models on the basis of approximation of unseen samples, and demonstrated the stability in the transfer of expression and rotation. 
From the results, we conclude that the proposed approach is a powerful way for characterizing and parameterizing the latent space of StyleGAN. 

% Future work
The current setting assumes complete data that contains measurements of all the people performing the same expressions from each rotation without any missing data. This requirement could be relaxed by low-rank completion methods that is left for future work.
To conclude we employed a model trained on FFHQ, and received promising results on the BU-3DFE data set. 
%
% We expect that using a model trained explicitly on the BU-3DFE will be even more beneficial for expression and rotation transfer that we will investigate in future. 